\documentclass[a4paper,twoside]{article}

\usepackage{hyperref}
\usepackage{epsfig}
\usepackage{subfigure}
\usepackage{calc}
\usepackage{amssymb}
\usepackage{amstext}
\usepackage{amsmath}
\usepackage{amsthm}
\usepackage{multicol}
\usepackage{pslatex}
\usepackage{apalike}
\usepackage{SCITEPRESS}     

\usepackage{xcolor}

\graphicspath{{figures/}}

\begin{document}

\title{FootAndBall: Integrated player and ball detector}
\author{\authorname{Jacek Komorowski \sup{1, 2}, Grzegorz Kurzejamski \sup{1, 2} and Grzegorz Sarwas \sup{1, 2}}
\affiliation{\sup{1}Warsaw University of Technology, Warsaw, Poland}
\affiliation{\sup{2}Sport Algorithmics and Gaming Sp. z o.o., Warsaw, Poland}
\email{jacek.komorowski@pw.edu.pl, g.kurzejamski@sagsport.com, sarwasg@ee.pw.edu.pl}}


\abstract{The paper describes a deep neural network-based detector dedicated for ball and players detection in high resolution, long shot, video recordings of soccer matches.
The detector, dubbed \emph{FootAndBall}, has an efficient fully convolutional architecture and can operate on input video stream with an arbitrary resolution. 
It produces ball confidence map encoding the position of the detected ball, player confidence map and player bounding boxes tensor encoding players' positions and bounding boxes.
The network uses Feature Pyramid Network desing pattern, where lower level features with higher spatial resolution are combined with higher level features with bigger receptive field. This improves discriminability of small objects (the ball) as larger visual context around the object of interest is taken into account for the classification. 
Due to its specialized design, the network has two orders of magnitude less parameters than a generic deep neural network-based object detector, such as SSD or YOLO. This allows real-time processing of high resolution input video stream. Our code and pre-trained model can be found on the project website: \url{https://github.com/jac99/FootAndBall}.
}

\onecolumn \maketitle \normalsize \vfill

\section{\uppercase{Introduction}}
\label{sec:introduction}
\noindent 


Accurate and efficient ball and player detection is a key element of any solution intended to 
automate analysis of video recordings of soccer games.
The method proposed in this paper allows effective and efficient ball and player detection in long shot, high definition, video recordings.
It's intended as a key component of the computer system developed for football academies and clubs to automate analysis of soccer video recordings. 

\begin{figure}
    \centering
    \includegraphics[trim={0 0.5cm 1.1cm 0},clip, width=0.9 
    \columnwidth]{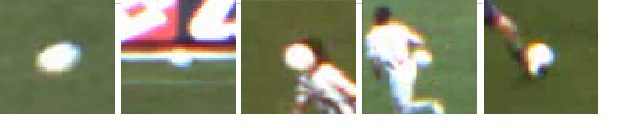}  \\
    \includegraphics[height=1.76cm]{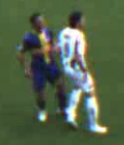}
    \includegraphics[height=1.76cm]{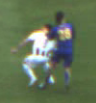}
    \includegraphics[height=1.76cm]{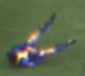}
    \includegraphics[height=1.76cm]{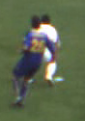}
    \caption{Exemplary patches illustrating difficulty of the ball and players detection task.
    Images of the ball exhibit high variance due to motion blur or can be occluded by a player. Players can be in a close contact and occluded.}
    \label{jk:fig:ball_examples}
\end{figure}

Detecting the ball from long-shot video footage of a soccer game is a challenging problem. 
\cite{deepball} lists the following factors that make the problem of ball localization difficult.
First, the ball is very small compared to other objects visible in the observer scene. Its size varies significantly depending on the position. In long shot recordings of soccer games, the ball can appear as small as 8 pixels, when it's on the far side of the pitch, opposite from the camera; and as big as 20 pixels, when it's on the near side of the field. At such small size, the ball can appear indistinguishable from parts of the players body (e.g. head or white socks) or the background clutter, such as small litter on the pitch or parts of stadium advertisements.
The shape of the ball can vary. When it's kicked and moves at high velocity, it becomes blurry and elliptical rather then circular. Perceived colour changes due to shadows and lighting variation. Situations when the ball is in player's possession or partially occluded are especially difficult. Simple ball detection methods based on motion-based background subtraction fail in such cases.
Top row of Fig.~\ref{jk:fig:ball_examples} shows exemplary image patches illustrating variance in the ball appearance and difficulty of the ball detection task. 

Players are larger than a ball and usually easier to detect. But it some situations their detection can be problematic. 
Players are sometimes in close contact with each other and partially occluded. They can have unusual pose due to stumbling and falling on the pitch. See bottom row of Fig.~\ref{jk:fig:ball_examples} for exemplary images showing difficulty of the player detection task.

In this paper we present a ball and players detection method inspired by recent progress in deep neural network-based object detection methods. 
Our method operates on a single video frame and is intended as the first stage in the soccer video analysis pipeline.
The detection network is designed with performance in mind to allow efficient processing of high definition video. 
Compared to the generic deep neural-network based object detector it has two orders of magnitude less parameters (e.g. 120 thousand parameters in our model versus 24 million parameters in SSD300~\cite{Liu16}).
Evaluation results in Section \ref{jk:ev_results} prove that our method can efficiently process high definition video input in a real time. It achieves 37 frames per second throughput for high resolution (1920x1080) videos using low-end GeForce GTX 1060 GPU. 
In contrast, Faster-RCNN~\cite{Ren15} generic object detector runs with only 8 FPS on the same machine.


\section{\uppercase{Related Work}}
\label{sec:related_work}
\noindent

The first step in traditional ball or player detection methods, is usually the background subtraction. 
The most commonly used background subtraction approaches are based on chromatic features~\cite{Gong95,Yoon02} 
or motion-based techniques such as background subtraction~\cite{DOr02,Mazz12}.
Segmentation methods based on chromatic features use domain knowledge about the visible scene: football pitch is mostly green and the ball mostly white.  
The colour of the pitch is usually modelled using a Gaussian Mixture Model and hardcoded in the system or learned. When the video comes from the static camera, motion-based segmentation is often used. 
For computational performance reasons, a simple approach is usually applied based on an absolute difference between consecutive frames or the difference between the current frame and the mean or median image obtained from a few previously processed frames \cite{High16}.

Traditional ball detection methods often use heuristic criteria based on chromatic or morphological features of connected components obtained from background segmentation process.
These criteria include blob size, colour and shape (circularity, eccentricity). Variants of Circle Hough Transform, modified to detect elliptical rather than circular objects, are used to verify if a blob contains the ball ~\cite{DOr02,Popp10,Halb15}.
To achieve real-time performance and high detection accuracy, 
a two-stage approach may be employed~\cite{DOr02}. First, regions that probably contain the ball are found (\emph{ball candidates extraction}), then candidates are validated (\emph{ball candidate validation}).

Ball detection methods using morphological features to analyze shape of blobs produced by background segmentation, fail if a ball is touching a player. 
See three rightmost examples in the top row of Fig.~\ref{jk:fig:ball_examples} for situations where these methods are likely to fail. 

Earlier player detection methods are based on connected component analysis~\cite{Yoon02,Dor07} or use techniques such as variants of Viola and Jones detector with Haar features~\cite{Lehu07,Liu09}.
\cite{Lehu07} present a football player detection method based on convolutional neural networks. The network has relatively shallow architecture and produces feature maps indicating probable player positions. 
\cite{Mack10} uses a combination of Histogram of Oriented Gradients (HOG) and Support Vector Machines (SVM) for player detection. First background segmentation, using a domain football pitch colour, is performed. HOG descriptors are classified by linear SVM algorithm trained on player template database.

\cite{Lu13} uses Deformable Part Model (DPM)~\cite{Felz13} to detect sport players in video frames. 
The DPM consists of 6 parts and 3 aspect ratios. 
The weakness of this method is that it may fail to detect partially occluded players due to non-maximum suppression operator applied after detection.

A comprehensive review of player detection and tracking methods in~\cite{Mana17} lists the following weaknesses present in reviewed algorithms. 
One of the most frequent is the problem with correctly detecting partially occluded players.
Motion-based techniques using background subtraction fail if a player is standing still or moving very slowly for a prolonged time.
Discrimination of players wearing white jerseys from the pitch white lines can be challenging using colour cues. 
Pixel-based player detection methods often fragment player into multiple separated regions (e.g. parts of legs) due to variations in the colour of legs, jerseys, shorts and socks.
Template-based detectors have problems with handling occlusions or situations when a players has an unusual pose (e.g. due to stumbling and falling on the pitch). 
See bottom row of Fig.~\ref{jk:fig:ball_examples} for exemplary difficult situations for the player detection task.

In recent years a spectacular progress was made in the area of neural-network based object detection methods.
Deep neural-network based YOLO detector~\cite{Redm16} achieves 63\% mean Average Precision (mAP) on PASCAL VOC 2007 dataset, whereas traditional Deformable  Parts  Models (DPM) detector~\cite{Felz10} scores only 30\%. Current state-of-the-art object detectors can be categorized as one-stage or two-stage. In two-stage detector, such as: Fast R-CNN~\cite{Girs15} or Faster R-CNN~\cite{Ren15},
the first stage generates a sparse set of candidate object locations (region proposals). The second stage uses deep convolutional neural network to classify each candidate location as one of the foreground classes or as a background. One-stage detectors, SSD~\cite{Liu16} or YOLO~\cite{Redm16}, do not include a separate region-proposal generation step. A single detector based on deep convolutional neural network is applied instead. 
However general-purpose neural network-based object detectors are relatively large scale networks with tens of millions of trainable parameters (e.g. 24 million parameters in SSD300~\cite{Liu16} detector). Another drawback is their limited performance on on small object detection. Authors of SSD method report significant drop of performance when detecting small objects. On COCOtest-dev2015 dataset average precision drops from 41.9\% for large object to only 9.0\% for small objects. 
This restricts usage of generic object detectors to recognize small objects, such as the ball.

Following the successful applications of convolutional neural networks to solve many computer vision problems, a few neural network-based ball and player detection methods were recently proposed.

\cite{Spec17} uses convolutional neural networks (CNN) to localize the ball under varying environmental conditions. The first part of the network consists of multiple convolution and max-pooling layers which are trained on the standard object classification task. The output of this part is processed by fully connected layers regressing the ball location as probability distribution along x- and y-axis. The network is trained on a large dataset of images with annotated ground truth ball position. 
The limitation of this method is that it fails if more than one ball, or object very similar to the ball, is present in the image.
\cite{Reno18} presents a 
deep neural network classifier, consisting of convolutional feature extraction layers followed by fully connected classification layer. It is trained to classify small, rectangular image patches as ball or no-ball.
The classifier is used in a sliding window manner to generate a probability map of the ball occurrence. 
The method has two drawbacks. 
First, the set of negative training examples (patches without the ball) must be carefully chosen to include sufficiently hard examples. 
Also the rectangular patch size must be manually selected to take into account all the possible ways the ball appears on the scene: big or small due to the perspective, sharp or blurred due to its speed. The method is also not optimal from the performance perspective.
Each rectangular image patch is separately processed by the neural network using a sliding-window approach. Then, individual results are combined to produce a final ball probability map.
\cite{gabel2018jetson} uses off-the-shelf deep neural network-based classifier architectures (AlexNet and Inception) fine tuned to detect the ball in rectangular patches cropped from the input image.
Authors reported 99\% ball detection accuracy on the custom dataset from RoboCup competition.
However, their dataset contains much closer, zoomed, views of the pitch, with the ball size considerably larger than in a long shot videos used as an input to our method. Also off-the-shelf architectures, such as Inception, are powerful but require much more computational resources for training and inference .
\cite{kamble2019deep} describes deep learning approach for 2D ball and player detection and tracking in soccer videos.
First, median filtering-based background subtraction is used to detect moving objects.
Then, extracted patches are classified into three categories: ball, player and background using VGG-based classifier.
Such approach fails if ball is touching or partially occluded by the player.
\cite{csah2018evaluation} uses a similar approach for player detection, where fixed-size images patches are extracted from an input image using a sliding window approach. The patches are initially filtered using hand crafted rules and then fed into a classification CNN with a relatively shallow, feed forward architecture. 

In contract to above patch-based methods, our solution requires a single pass of an entire image through the network. It can be efficiently implemented using modern deep learning frameworks and use the full capabilities of GPUs hardware.
It does not require extraction, resizing and processing of multiple separate patches.

\cite{Lu17} presents a cascaded convolutional neural network (CNN) for player detection. The network is lightweight and thanks to cascaded architecture the inference is efficient. The training consists of two phases: branch-level training and whole network training and cascade thresholds are found using the grid search.
Our method is end-to-end trainable in a single phase and requires less hyper-parameters as multiple cascade thresholds are not needed. Training is also more efficient as our network processes entire images in a single pass, without the need to extract multiple patches. 

\cite{zhang2018rc} is a player detection method based on SSD~\cite{Liu16} object detector. Authors 
highlight the importance of integrating low level and high level (semantic) features to improve detection accuracy. To this end, they introduce reverse connected modules which integrate features derived from multiple layers into multi-scale features. In principle, the presented design modification is similar to Feature Pyramid Network~\cite{Lin17} concept, where higher level, semantically richer features are integrated with lower level features.

DeepBall~\cite{deepball} method is an efficient ball detection method based on fully convolutional architecture. It takes an input image and produces a ball confidence map indicating the most probable ball locations. 

The method presented in this paper proposes a unified solution to efficiently detect both players and the ball in input images. It's architectured to effectively detect objects of different scales such as the ball and players by using Feature Pyramid Network~\cite{Lin17} design pattern.

\section{\uppercase{Player and ball detection method}}
\label{sec:method}
\noindent

The method presented in this paper, dubbed \emph{FootAndBall}, is inspired by recent developments of neural network-based generic object detectors, such as SSD~\cite{Liu16} and \emph{DeepBall}~\cite{deepball} ball detection method.
Typical architecture of a single-stage, neural network-based, object detector is modified, to make it more appropriate for player and ball detection in long shot soccer videos.
Modifications aim at improving the processing frame rate by reducing the complexity of the underlying feature extraction network. 
For comparison, SSD300~\cite{Liu16} model has about 24 million parameters, while our model only 199 thousand, more than 200 times less
Multiple anchor boxes, with different sizes and aspect ratios, are not needed as we detect objects from two classes with a limited shape and size variance.
The feature extraction network is architectured to improve detection accuracy of object with different scales: small objects such as the ball and larger players.
It's achieved by using Feature Pyramid Network~\cite{Lin17} design pattern which efficiently combines higher resolution, low level feature maps with lower resolution maps encoding higher-level features and having larger receptive field.
This allows both precise ball localization and improved detection accuracy in difficult situations as larger visual context allows differentiating the ball from parts of player body and background clutter (e.g. parts of stadium advertisement).

Generic single-shot object detection methods usually divide an image into the relatively coarse size grid (e.g. 7x7 grid in YOLO~\cite{Redm16}) and detect no more than one object of interest with particular aspect ratio in each grid cell. 
This prevents correct detections when two objects, such as players, are close to each other. Our methods uses denser grids.
For input image of 1920x1080 pixels we use 480x270 grid (input image size scaled down by 4) for ball detection and 120x68 grid (input image size scaled down by 16) for player detection. This allows detecting two near players as two separate objects.

\begin{figure*}
    \centering
    \includegraphics[clip, trim=1cm 18.0cm 1.9cm 1.0cm, width=0.95\textwidth]{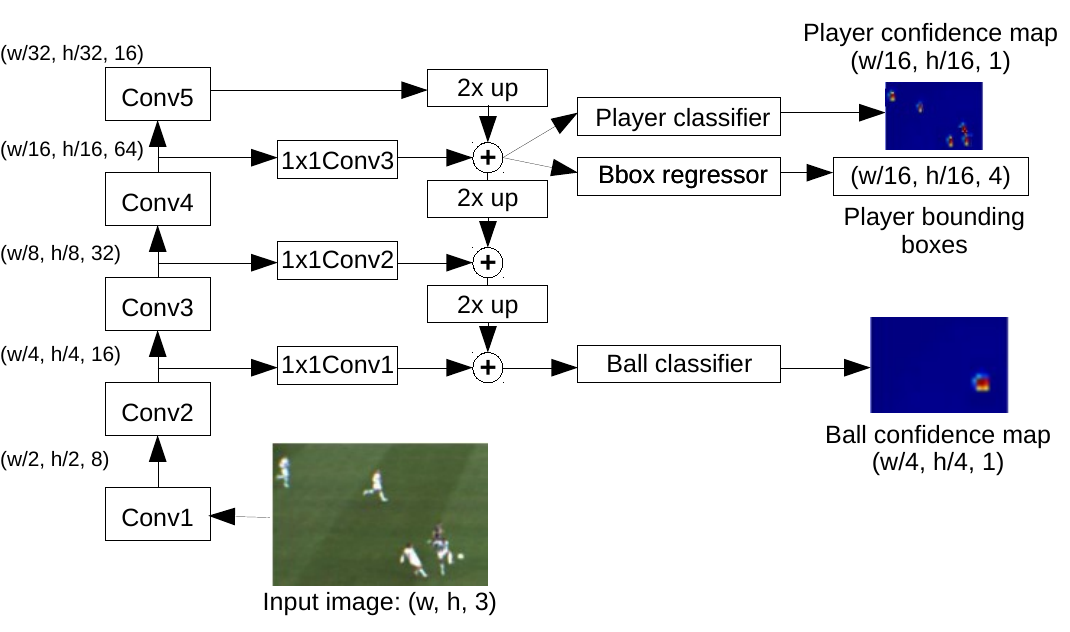}
    \caption{High-level architecture of \emph{FootAndBall} detector.
    The input image is processed bottom-up by five convolutional blocks (Conv1, Conv2, Conv3, Conv4 and Conv5) producing feature maps with decreasing spatial resolution and increasing number of channels.
    The feature maps are then processed in the top-down direction. Upsampled feature map from the higher pyramid level is added to the  feature map from the lower level. 1x1 convolution blocks decrease the number of channels to the same value (32 in our implementation).
    Resultant feature maps are processed by three heads: ball classification head, player classification head and player bounding box regression head. 
    Numbers in brackets denote size of feature maps (width, height, number of channels) produced by each block, where \(w, h\) is the input image width and height.
    }
    \label{jk:fig:network-diagram}
\end{figure*}

Our method takes an input video frame and produces three outputs:
\emph{ball confidence map} encoding probability of ball presence at each grid cell,
\emph{player confidence map} encoding probability of player presence at each grid cell
and \emph{player bounding box tensor} encoding coordinates of a player bounding box at each cell of the player confidence map.
See Fig.~\ref{jk:fig:network-diagram} for visualization of network outputs.

For the input image with \(w \times h\) resolution, 
the size of the output \emph{ball confidence map} is \(w/k_B \times h/k_B\), where \(k_B \) is the scaling factor (\(k_B=4\) in our implementation). 
Position in the \emph{ball confidence map} with coordinates \((i, j)\) corresponds to the position \((\lfloor k_B(i-0.5) \rfloor, \lfloor k_B(j-0.5) \rfloor\) in the input image.
The ball is located by finding the location in the ball confidence map with the highest confidence.
If the highest confidence is lower than a threshold  \(\Theta_{B}\), no balls are detected. This can happen when a ball is occluded by a player or outside the pitch.
The pixel coordinates of the ball in the input frame are computed using the following formula:
\((x, y) = (\lfloor k_B(i-0.5) \rfloor, \lfloor k_B(j-0.5) \rfloor\), 
where \((i, j)\) are coordinates in the \emph{ball confidence map}.

The size of the \emph{player confidence map} is  \(w/k_P \times h/k_P\), where \(k_P\) is the scaling factor (\(k_P=16\) in our implementation).
The size of the \emph{player bounding box tensor} is \(w/k_P \times h/k_P \times 4\).
The tensor encodes four bounding box coordinates for each location in the player confidence map where the player is detected. 
Position in the \emph{player confidence map} with coordinates \((i, j)\) corresponds to the position \((\lfloor k_P(i-0.5) \rfloor, \lfloor k_P(j-0.5) \rfloor\) in the input image.
Player positions are found by first finding finding local maxima in the player confidence map with the confidence above the threshold \(\Theta_{P}\). This is achieved by applying non maximum suppression to the player confidence map and taking all locations with the confidence above the threshold \(\Theta_{P}\).
Let \(P = \left\{ \left( i,j \right) \right\}\) be the set of such local maxima.
For each identified player location \((i,j)\) in the player confidence map, we retrieve bounding box coordinates from the bounding box tensor. 
Similar to SSD~\cite{Liu16}, player bounding boxes are encoded as \((x_{bbox}, y_{bbox}, w_{bbox}, h_{bbox}) \in \mathbb{R}^4\) vectors, where \((x_{bbox}, y_{bbox})\) is a relative position of the centre of the bounding box with respect to the center of the corresponding grid cell in the player confidence map and \(w_{bbox}\), \(h_{bbox}\) are its width and height. The coordinates are normalized to be in $0..1$ scale.
The bounding box centre in pixel coordinates is calculated as: 
\((x_{bbox}', y_{bbox}') = (\lfloor k_P(i-0.5) + x_{bbox} w \rfloor, \lfloor k_P(j-0.5) + y_{bbox} h \rfloor)\), where \( (w, h) \) is an input image resolution.
It's height and width in pixel coordinates is \((\lfloor w_{bbox} w\rfloor, \lfloor h_{bbox} h \rfloor)\).

\begin{figure}
    \centering
    \includegraphics[width=0.47\textwidth]{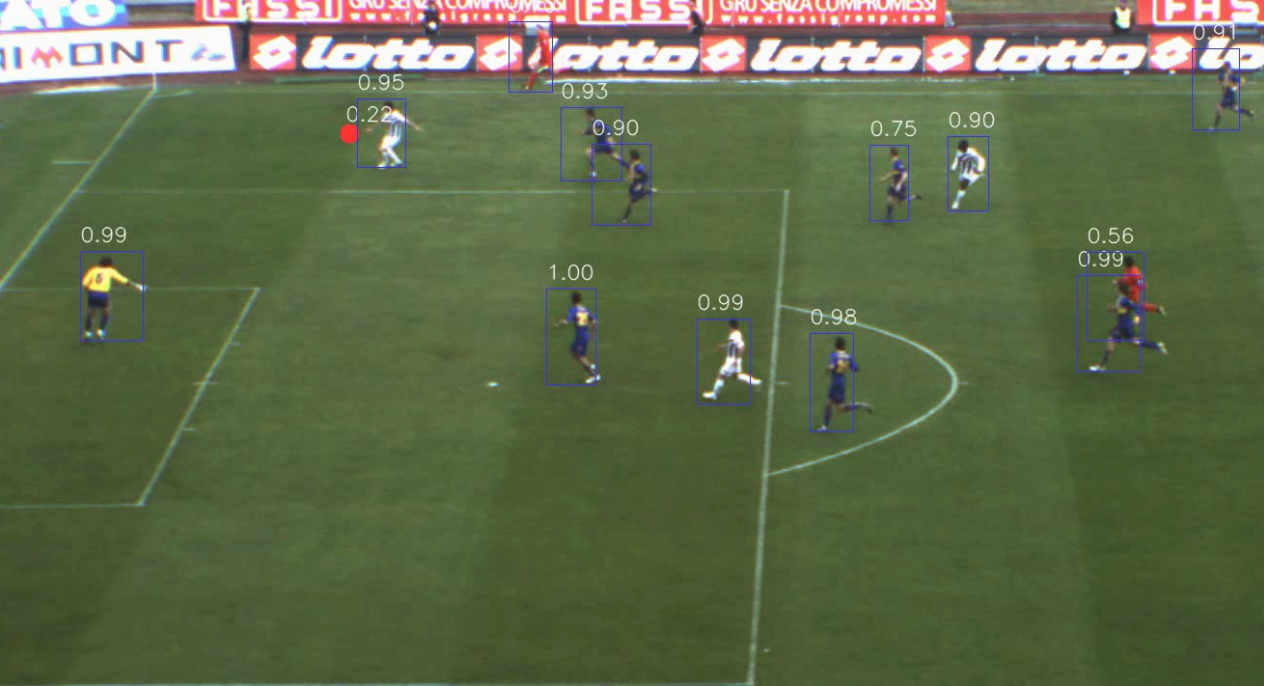}
    \includegraphics[width=0.47\textwidth]{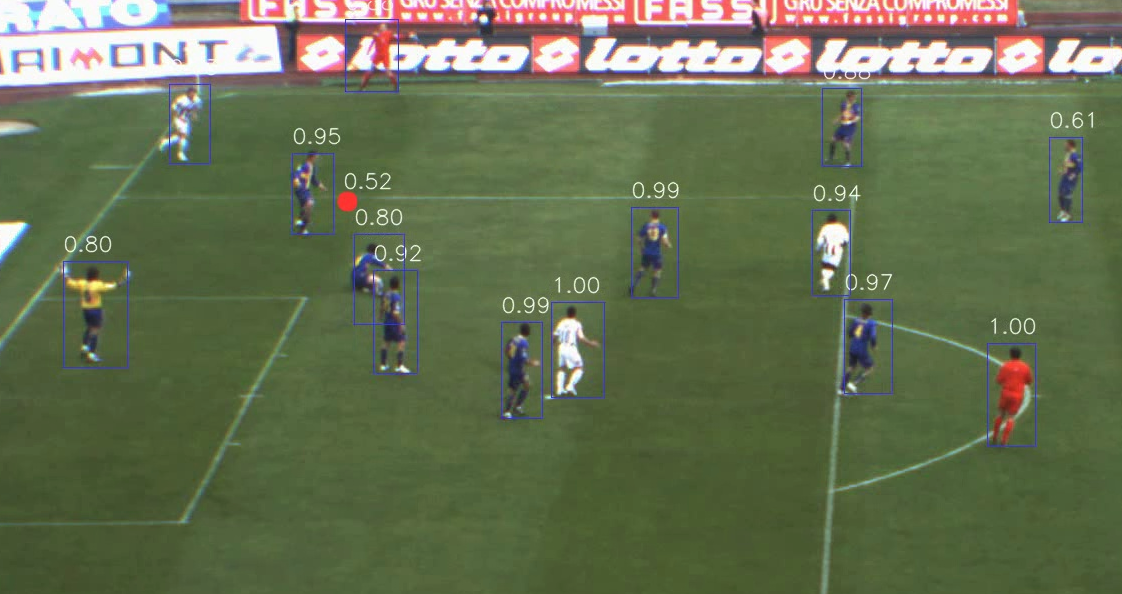}
    \caption{Player and ball detection results. Numbers above bounding boxes show detection confidence from the player classifier.}
    \label{jk:fig:det_results}
\end{figure}

\begin{figure}
    \centering
    \includegraphics[height=2cm]{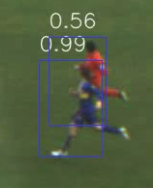}
    \includegraphics[height=2cm]{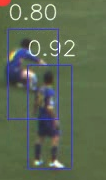}    
    \includegraphics[height=2cm]{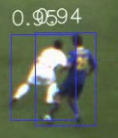}   
    \includegraphics[height=2cm]{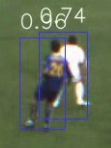}
    \includegraphics[height=2cm]{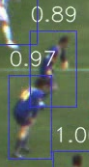}
    \caption{Detection results in difficult situations (player occlussion).}
    \label{jk:fig:det_overlap}
\end{figure}

Detection results are visualized in Fig.~\ref{jk:fig:det_results}. Numbers above the player bounding boxes show detection confidence. Detected ball position is indicated by the red circle.
Fig.~\ref{jk:fig:det_overlap} show player detection performance in difficult situations, when players' are touching each other or occluded. 

\paragraph{Network architecture}

\begin{table}
	\centering
    \begin{tabular}{lll}
     \hline
    Block  &  Layers &  Output size\\
     \hline
    Conv1  &  16 filters 3x3   \\
           & Max pool 2x2 & $(w/2, h/2, 16)$\\
    Conv2  &  32 filters 3x3   \\
           & 32 filters 3x3   \\
           & Max pool 2x2  & $(w/4, h/4, 32)$\\
    Conv3  &  32 filters 3x3    \\
           & 32 filters 3x3  \\
           & Max pool: 2x2  & $(w/8, h/8, 32)$\\
    Conv4  &  64 filters 3x3   \\
           & 64 filters 3x3  \\
           & Max pool: 2x2  & $(w/16, h/16, 64)$ \\
    Conv5  &  64 filters 3x3    \\
           & 64 filters 3x3  \\
           & Max pool 2x2  & $(w/32, h/32, 32)$ \\
           
    \hline
    1x1Conv1 &  32 filters 1x1  & $(w/4, h/4, 32)$\\
    1x1Conv2 &  32 filters 1x1  & $(w/8, h/8, 32)$\\
    1x1Conv3 &  32 filters 1x1  & $(w/16, h/16, 32)$\\
    \hline
    Ball   & 32 filters 3x3  &  \\
    classifier                 & 2 filters 3x3 & \\
                     & Sigmoid &  $(w/4, h/4, 1)$ \\
   Player   &  32 filters 3x3 \\
    classifier  & 2 filters 3x3  \\
                     & Sigmoid &  $(w/16, h/16, 1)$ \\
    BBox    &  32 filters 3x3 \\
    regressor  & 4 filters 3x3 & $(w/16, h/16, 4)$ \\
    \hline
	\end{tabular}
	\caption{Details of \emph{FootAndBall} detector architecture. Third column lists size of the output from each block as (width, height, number of channels) for an input image with \(w \times h\) resolution.
	All convolutional layers, except for 1x1 convolutions, are followed by BatchNorm and ReLU non-linearity (not listed in the table for brevity). All convolutions use 'same' padding and stride one.
    }
    \label{jk:tab-details}
\end{table}

Figure~\ref{jk:fig:network-diagram} shows high level architecture of our \emph{FootAndBall} network.  It's based on Feature Pyramid Network~\cite{Lin17} design pattern.
The input image is processed in the bottom-up direction by five convolutional blocks (Conv1, Conv2, Conv3, Conv4, Conv5) producing feature maps with decreasing spatial resolution and increasing number of channels. 
Numbers in brackets denote the width, height and number of channels \((w, h, c)\) of feature maps produced by each block.
The feature maps are then processed in the top-down direction. Upsampled feature maps from the higher pyramid level are added to feature maps from the lower level. 
1x1 convolution blocks decrease the number of channels to the same value (16 in our implementation), so two feature maps can be summed up.
Resultant feature maps are processed by three heads: ball classification head, player classification head and player bounding box regression head. 
Ball classification head takes a feature map with spatial resolution \(w/4 \times h/4\) as an input and produces one channel \(w/4 \times h/4\) \emph{ball confidence map} denoting probable ball locations. Each location in the ball confidence map corresponds to \(4 \times 4\) pixel block in the input image.
Player classification head takes a feature map with spatial resolution \(w/16 \times h/16\) as an input and produces one channel \(w/16 \times h/16\) \emph{player confidence map} denoting detected player locations.
Each location in the player confidence map corresponds to \(16 \times 16\) pixel block in the input image. Player bounding box regression head takes the same input as the player classification head.
It produces produces four channel \(w/16 \times h/16 \times 4\) \emph{player bounding box tensor} encoding coordinates of a player bounding box at each location of the player confidence map.
Details of each block are listed in Table~\ref{jk:tab-details}.
The network has fully convolutional architecture and can operate on images of any size.

Using Feature Pyramid Network~\cite{Lin17} architecture allows using both low-level features from the first convolutional layers and high-level features computed by higher convolutional layers.
Information from first convolutional layers is necessary for a precise spatial location of the object of interest.
Further convolutional layers operate on feature maps with lower spatial resolution, thus they cannot provide exact spatial location. 
But they have bigger receptive fields and their output provides additional context to improve classification accuracy.
This is especially important for the ball detection tasks. The ball is very small and other objects, such as parts of players' body or stadium advertisement, may have similar appearance.
Using larger receptive field and higher level features can improve discriminability of the ball detector.

\paragraph{Loss function} is a modified version of the loss used in SSD~\cite{Liu16} detector.
The loss minimized during the neural network training consists of three components: ball classification loss, player classification loss and player bounding box loss. 
Proposed network does not regress ball bounding box. 
Ball classification loss (\(\mathcal{L}_{b}\)) is a binary cross-entropy loss over predicted ball confidence and the ground truth:
\begin{equation}
\mathcal{L}_{B} =
-\sum_{\left(i,j\right)\in Pos^{B}} \log c_{ij}^{B} 
-\sum_{\left(i,j\right)\in Neg^{B}} \log \left( 1 - c_{ij}^{B}  \right) ,
\end{equation}
where \(c_{ij}^{B}\) is the value of the ball confidence map at the spatial location \((i,j)\).
\(Pos^{B}\) is a set of positive ball examples, that is the set of locations in the ball confidence map corresponding to the ground truth ball position (usually for one input image it's only one location). 
\(Neg^{B}\) is a set of negative examples, that is the set of locations that does not correspond to the ground truth ball position (all locations do not containing the ball).
Player classification loss (\(\mathcal{L}_{p}\)) is a binary cross-entropy loss over predicted confidence in each cell in the player confidence map and the ground truth.
\begin{equation}
\mathcal{L}_{p} =
-\sum_{\left(i,j\right)\in Pos^{P}} \log c_{ij}^{P} 
-\sum_{\left(i,j\right)\in Neg^{P}} \log \left( 1 - c_{ij}^{P}  \right) ,
\end{equation}
where \(c_{ij}^{P}\) is the value of the player confidence map at the spatial location \((i,j)\).
\(Pos^{P}\) is a set of positive player examples, that is the set of locations in the player confidence map corresponding to the ground truth player position. 
\(Neg^{P}\) is a set of negative examples, that is the set of locations that does not correspond to any ground truth player position.
Player bounding box loss (\(\mathcal{L}_{bbox}\)) is Smooth L1 loss~\cite{girshick2015fast} between the predicted and ground truth bounding boxes.
As in SSD~\cite{Liu16} detector we regress the offset of the bounding box with respect to the cell center and its width and height.
\begin{equation}
\mathcal{L}_{bbox} =
\sum_{\left(i,j\right)\in Pos^{p}}
\mathrm{smooth_{L1}}
\left(
l_{(i,j)} - g_{(i,j)}
\right) ,
\end{equation}
where \(l_{(i,j)} \in \mathbb{R}^4\) denotes a predicted bounding box coordinates in the location \((i,j)\)
and \(g(i,j) \in \mathbb{R}^4\) are a ground truth bounding box coordinates in the location \((i,j)\).

Similar to SSD, we regress relative position of the center \((cx, cy)\) of the  bounding box with respect to the location centre and its relative width (\(w\)) and height (\(w\)) with respect to the location/cell height and width.

The total loss is the sum of the three above components, averaged by the number of training examples \(N\):
\begin{equation}
\mathcal{L} = \frac{1}{N}
\left(
\alpha_B \mathcal{L}_{B} + \alpha_B \mathcal{L}_{P} + \mathcal{L}_{bbox}
\right) ,
\end{equation} 
where \(\alpha_B\) and \(\beta_P\) are weights for the \(\mathcal{L}_{B}\)
and \(\mathcal{L}_{P}\) chosen experimentally.


Sets of positive ball \( Pos^{B} \) and player \( Pos^{P} \) examples  are constructed as follows. 
If \((x,y)\) is a ground true ball position in pixel coordinates, then the corresponding confidence map location \((i,j) = ( \lfloor x/k_B, y/k_B \rfloor )\) and all neighbourhood locations are added to \(Pos^{B}\).
If \((x,y)\) is a ground truth position of the player's bounding box center in pixel coordinates, then the corresponding confidence map location \((i,j) = ( \lfloor x/k_P, y/k_P \rfloor )\) is added to \(Pos^{P}\). 
Due to smaller size of players confidence map, we mark only a single cell as a positive example.

Sets of negative ball \( Neg^{B} \) and negative player \( Neg^{P} \) examples correspond to locations in the confidence map, where the objects are not present.
The number of negative examples is orders of magnitude higher than a number of positive examples and this would create highly imbalanced training set.
To mitigate this, we employ hard negative mining strategy as in~\cite{Liu16}. For both ball and players we chose a limited number of negative examples with the highest confidence loss, so the ratio of negative to positive examples is at most 3:1.

\paragraph{Network training}
The network is trained using combination of two datasets: ISSIA-CNR Soccer~\cite{DOr09} and Soccer Player Detection~\cite{Lu17} datasets.
ISSIA-CNR Soccer dataset contains six synchronized, long shot views of the football pitch acquired by six Full-HD DALSA 25-2M30 cameras. Three cameras are designated for each side of the playing-field, recording at 25 fps. Videos are acquired during matches of the Italian 'serie A'. There're 20,000 annotated frames in the dataset annotated with ball position and player bounding boxes. Fig. \ref{jk:fig:det_results} shows exemplary frames from the ISSIA-CNR Soccer Dataset.
Soccer Player Detection dataset is created from two professional football matches.  Each match was recorded by three broadcast cameras at 30 FPS with 1280×720 resolution.  It contains 2019 images with 22,586 annotated player locations. However, ball position is not annotated.

For training we select 80\% of images and use remaining 20\% for evaluation. We use data augmentation to increase the variety of training examples and reduce overfitting. The following transformations are randomly applied to the training images: random scaling (with scale factor between 0.8 and 1.2), random cropping, horizontal flip and random photometric distortions (random change of brightness, contrast, saturation or hue). The ground truth ball position and player bounding boxes are modified accordingly to align with the transformed image.

The network is trained using a standard gradient descent approach with Adam~\cite{King14} optimizer. The initial learning rate is set to \(0.001\) and decreased by 10 after 75 epochs. The training runs for 100 epochs in total. Batch size is set to 16. 

\section{\uppercase{Experimental results}}
\label{jk:section-experimental-results}
\noindent

\paragraph{Evaluation dataset}
We evaluate our method on publicly available ISSIA-CNR Soccer~\cite{DOr09} and Soccer Player Detection~\cite{Lu17} datasets. 
For evaluation we select 20\% of images, whereas remaining 80\% are used for training.
Both datasets are quite challenging, there's noticeable motion blur, many occlusions, cluttered background and varying player's size.
In ISSIA-CNR dataset the height of players is between 63 and 144 pixels. In Soccer Player Detection dataset, it varies even more, from 20 to 250 pixels.
In ISSIA-CNR dataset one team wears white jerseys which makes difficult to  distinguish the ball when it's close to the player.

\paragraph{Evaluation metrics}
\label{sec:metrics}
We evaluate Average Precision (AP), a standard metric used in assessment of object detection methods. We follow Average Precision definition from Pascal 2007 VOC Challenge \cite{Ever10}.
The precision/recall curve is computed from a method’s ranked output. 
For the ball detection task, the set of positive detections is the set of ball detections (locations in the \emph{ball confidence map} with the confidence above the threshold \(\Theta_B\)) corresponding to the ground truth ball position. Our method does not regress bounding boxes for the ball.
For the player detection task, the set of positive detections is the set of player detections (estimated player bounding boxes) with Intersection over Union (IOU) with the ground truth players' bounding boxes above 0.5. 
Recall is defined as a proportion of all positive detections with confidence above a given threshold to all positive examples in the ground truth. 
Precision is a proportion of all positive detections with confidence above that threshold to all examples. Average Precision (AP) summarizes the shape of the precision/recall curve, and is defined as the mean precision at a set of eleven equally spaced recall levels:

\begin{equation}
\mathrm{AP} = \frac{1}{11} \sum_{r \in \left\{ 0, 0.1, \ldots 1 \right\}} p(r) \; ,\end{equation} 
where \(p(r)\) is a precision at recall level \(r\).

\paragraph{Evaluation results}
\label{jk:ev_results}
Evaluation results are summarized in Table~\ref{jk:table1}. The results show Average Precision (AP) for ball and player detection as defined in the previous section. 
The table also lists a number of parameters of each evaluated model and inference time, expressed in frames per second (FPS) achieved when processing Full HD (1920x1080 resolution) video.
All methods are implemented in PyTorch~\cite{Pasz17} 1.2 and run on low-end nVidia GeForce GTX 1060 GPU.

\begin{table*}[t]
\caption{Average precision (AP) of ball and player detection methods on ISSIA-CNR and Soccer Player datasets. Two last columns show the number of trainable parameters and inference time in frame per seconds for high resolution (1920, 1080) video. }
\begin{center}
\begin{tabular}{l@{\quad}c@{\quad}c@{\quad}c@{\quad}c@{\quad}c@{\quad}r@{\quad}r@{\quad}l}
\hline
Dataset & \multicolumn{3}{c}{ISSIA-CNR} & Soccer Player \\
 & \begin{tabular}{@{}c@{}}Ball AP \end{tabular}
& \begin{tabular}{@{}c@{}}Player AP \end{tabular}
& \begin{tabular}{@{}c@{}}mAP \end{tabular}
& \begin{tabular}{@{}c@{}}Player AP \end{tabular}
&\begin{tabular}{@{}c@{}}\#params\end{tabular}
&\begin{tabular}{@{}c@{}}FPS\end{tabular}
\\
\hline
\cite{Reno18} &  0.834 & - & - & - & 313k & 32 \\
DeepBall &  0.877 & - & - & - & 49k & 87 \\
\hline
Faster R-CNN & 0.866 & 0.874 & 0.870 & \textbf{0.928} & 25 600k & 8 \\
\hline
FootAndBall (no top-down) & 0.853 & 0.889 & 0.872 & 0.834 & 137k & 39 \\
\textbf{FootAndBall} & \textbf{0.909} & \textbf{0.921} & \textbf{0.915} & 0.885 & 199k & 37 \\
[2pt]
\hline
\end{tabular}
\end{center}
\label{jk:table1}
\end{table*}

Our method yields the best results on the ball detection task on ISSIA-CNR dataset. It achieves 0.909 Average Precision.
For comparison we evaluated three recent, neural network-based, detection methods.
Two are dedicated ball detection methods trained and evaluated on the same datasets: \cite{Reno18} scores 0.834 and \cite{deepball} 0.877 Average Precision. 
The other is general purpose Faster R-CNN~\cite{Girs15} object detector fine-tuned for ball and player detection using our training datasets. Despite having almost two order of magnitude more parameters is has lower ball detection average precision (0.866).

On the player detection task we compared our model to general purpose Faster R-CNN~\cite{Girs15} object detector. Comparison with other recently published  player detection method was not made due to reasons such as unavailability of the source code, usage of proprietary datasets, lack details on train/test split or different evaluation metrics than used in our paper.
On ISSIA CNR dataset our method achieves higher ball and player detection average accuracy (0.909 and 0.921 respectively) than Faster R-CNN (0.866 and 0.774) despite having two orders of magnitude less parameters.
On Soccer Player Dataset Faster R-CNN gets higher score (0.928 as compared to 0.885 scored by our method).
On average these two methods have similar accuracy. But due to the specialized design our method can process high definition video (1920 x 1080 frames) in a real time at 37 FPS. Faster R-CNN throughput on the same video is only 8 FPS (Faster-RCNN implementation based on ResNet-50 backbone, included in PyTorch 1.2 distribution \url{https://pytorch.org}).
DeepBall network has the highest performance (87 FPS) but it's a tiny network built for ball detection only.


\begin{figure}
  \centering
  \includegraphics[width=0.14\textwidth]{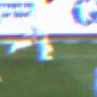} 
  \includegraphics[width=0.14\textwidth]{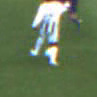}
  \includegraphics[width=0.14\textwidth]{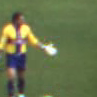} \\
  \includegraphics[width=0.14\textwidth]{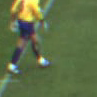}
  \includegraphics[width=0.14\textwidth]{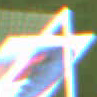}
  \includegraphics[width=0.14\textwidth]{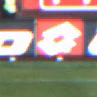}
  \caption{Visualization of incorrect ball detection results. Top row show image patches where the ball is not detected (false negatives). The bottom row shows patches with incorrectly detected ball (false positives).}
  \label{jk:fig:misclassifications}
\end{figure}


Using Feature Pyramid Network design pattern, where higher level features with larger receptive field are combined with lower level features with higher spatial resolution, is an important element of our design.
Evaluation of a similar network but without top-down connections and without combining multiple feature maps with different receptive fields, produces worse results. Such architecture (FootAndBall -- no top-down) achieves 4-5\% percentage points lower Average Precision in all categories.

Fig.~\ref{jk:fig:misclassifications} show examples of incorrect ball detections. Two top rows show image patches where our method fails to detect the ball (false negatives). It can be noticed, that misclassification is caused by severe occlusion, where only small part of the ball is visible, or due to blending of the ball image with white parts of the player wear or white background objects outside the play field, such as stadium advertisement. The bottom row shows examples of patches where a ball is incorrectly detected (false positives). The detector is sometimes confused by players' white socks or by the background clutter outside the play field.

\section{\uppercase{Conclusions}}
\noindent
The article proposes an efficient deep neural network-based player and ball detection method. 
The proposed network has a fully convolutional architecture processing entire image in a single pass through the network. This is much more computationally effective than a sliding window approach proposed in other methods, such as~\cite{Reno18}. Additionally, the network can operate on images of any size that can differ from size of images used during the training. It outputs ball locations and player bounding boxes. 
The method performs very well on a challenging ISSIA-CNR Soccer~\cite{DOr09} and Soccer Player Detection~\cite{Lu17} datasets (0.915 and 0.885 mean Average Precision).
In ball detection task it outperforms two other, recently proposed, neural network-based ball detections methods: \cite{Reno18} and \cite{deepball}.
In player detection task it's on par with fine-tuned general-purpose Faster R-CNN object detector, but due to specialized design it's almost five times faster (37 versus 8 FPS for high-definition 1920x1080 video). This allows real time processing of high-definition video recordings.

In the future we plan to use temporal information to increase the accuracy. Combining convolutional feature maps from subsequent frames may help to discriminate between object of interest and static distractors (e.g. parts of stadium advertisement or circular marks on the pitch).

Another research direction is to investigate model compression techniques, such as using half-precision arithmetic, to improve the computational efficiency. The proposed method can process high definition videos in real time on relatively low-end GPU platform. However, real time processing of recordings from multiple cameras poses a challenging problem.

\section*{Acknowledgements}
This work was co-financed by the European Union within the European Regional Development Fund  


\bibliographystyle{apalike}
{\small
\bibliography{db-bib}}

\vfill
\end{document}